# Identifying Bengali Multiword Expressions using Semantic Clustering

Tanmoy Chakraborty
*Department of Computer Science & Engineering, Indian Institute of Technology, Kharagpur, India – 721302*

Dipankar Das
*Department of Computer Science & Engineering, National Institute of Technology, Meghalaya, India*

Sivaji Bandyopadhyay
*Department of Computer Science & Engineering, Jadavpur University, Kolkata, India*

**Abstract:** One of the key issues in both natural language understanding and generation is the appropriate processing of Multiword Expressions (MWEs). MWEs pose a huge problem to the precise language processing due to their idiosyncratic nature and diversity in lexical, syntactical and semantic properties. The semantics of a MWE cannot be expressed after combining the semantics of its constituents. Therefore, the formalism of semantic clustering is often viewed as an instrument for extracting MWEs especially for resource constraint languages like Bengali. The present semantic clustering approach contributes to locate clusters of the synonymous noun tokens present in the document. These clusters in turn help measure the similarity between the constituent words of a potentially candidate phrase using a vector space model and judge the suitability of this phrase to be a MWE. In this experiment, we apply the semantic clustering approach for noun-noun bigram MWEs, though it can be extended to any types of MWEs. In parallel, the well known statistical models, namely Point-wise Mutual Information (PMI), Log Likelihood Ratio (LLR), Significance function are also employed to extract MWEs from the Bengali corpus. The comparative evaluation shows that the semantic clustering approach outperforms all other competing statistical models. As a byproduct of this experiment, we have started developing a standard lexicon in Bengali that serves as a productive Bengali linguistic thesaurus.

**Introduction**

Over the past two decades or so, Multiword Expressions (MWEs) have been identified with an increasing amount of interest in the field of computational linguistics and Natural Language Processing (NLP) (Rayson, 2010, p. 1). The term "MWE" is used to refer to various types of linguistic units and expressions including idioms (kick the bucket, 'to die'), compound noun (village community), phrasal verbs (find out, 'search'), other habitual collocations like conjunctions (as well as), institutionalized phrases (many thanks) etc. However, while there is no universally agreed definition for MWE as yet, most researchers use the term to refer to those frequently occurring phrasal units which are subject to a certain level of semantic opaqueness, or non-compositionality. Sag et al. (2002, p. 1) defined them as "idiosyncratic interpretations that cross word boundaries (or spaces)."

MWEs are treated as a special case in semantics since individual components of an expression often fail to keep their meanings intact within the actual meaning of that expression. This opaqueness in meaning may be partial or total depending on the degree of compositionality of the whole expression (Chakraborty, 2011, p. 8). MWEs have been studied for decades in Phraseology under the term "*phraseological unit*" (Baldwin, 2010, p. 267). But in the early 1990s, MWEs started receiving increasing attention in corpus-based computational linguistics and NLP. A number of research activities on MWEs have been carried out in various languages like English, German and many other European languages. Various statistical co-occurrence measurements like Mutual Information (MI) (Church, 1990, p. 22), Log-Likelihood (Dunning, 1993, p. 61), Salience (Kilgarriff and Rosenzweig, 2000, p. 15) have been suggested for the identification of MWEs. An unsupervised graph-based algorithm to detect compositionality of MWEs was also proposed in the research of Korkontzelos and Manandhar (2009, p. 65). On the other hand, Butnariu et al. (2009, p. 100) proposed a solution to interpret nominal compounds in English using paraphrasing and prepositions.

In the case of Indian languages, a considerable amount of research has been conducted in compound noun MWE extraction (Kunchukuttan and Damani, 2008, p. 20), complex predicate extraction (Das et al., 2010, p. 37), clustering based approach (Chakraborty et al., 2011, p. 8) and a classification based approach for identifying Noun-Verb collocations (Venkatapathy and Joshi, 2009, p. 899). Bengali, one of the more important Indo-Iranian languages, is the sixth-most popular language in the world and spoken by a population that now exceeds 250 million. Geographical Bengali-speaking population percentages are as follows: Bangladesh (over 95%), and the Indian States of Andaman & Nicobar Islands (26%), Assam (28%), Tripura (67%), and West Bengal (85%). The global total includes those which are spoken in the Diaspora in Canada, Malawi, Nepal, Pakistan, Saudi Arabia, Singapore, the United Arab Emirates, the United Kingdom, and the United States. In Bengali, works on automated extraction of MWEs are limited in number. One method of automatic extraction of Noun-Verb MWE in Bengali (Agarwal et al., 2004, p. 167) has been carried out using morphological evidence and significance function. They have classified Bengali MWEs based on the morpho-syntactic flexibilities and proposed a statistical approach for extracting the verbal compounds from a medium size corpus.

In this paper, we propose a framework for identifying MWEs from the perspective of semantic interpretation of MWEs that the meanings of the components are totally or partially diminished in order to construct the actual semantics of the expression. A clustering technique is employed to group all nouns that are related to the meaning of the individual component of an expression. Two types of similarity techniques based on vector space model are adapted to make a binary classification (MWE or Non-MWE) of potentially candidate phrases. We hypothesize that the more similar the components of an expression, the less probable that their combination forms a MWE. We test our hypothesis on the noun-noun bigram phrases. We also illustrate the efficiency of our model after translating the individual components of a phrase in English and fed these components into the WordNet::Similarity module module -- an open-source package developed at the University of Minnesota for calculating the lexical similarity between word (or sense) pairs based on variety of similarity measure. In this paper, we test our models with different cut-off values that define the threshold of (dis)similarity and the degree of compositionality of a candidate phrase. Experimental results corroborate our hypothesis that the dissimilarity of the meaning of constituent tokens enhances the chance of constructing a MWE. The use of English WordNet, quite strikingly, substantiates its enormous productivity in identifying MWEs from Bengali documents.

The remainder of this paper is organized as follows. Section 1 introduces a preliminary study about the Bengali MWEs and their morpho-syntactic based classification. Then the detailed description of candidate selection and the baseline system are described in section 2 and section 3 respectively. Section 4 illustrates traditional statistical methodologies for extracting MWEs from the document.

Section 5 presents an elaborate description of semantic clustering approach. The introduction of English WordNet::Similarity in identifying Bengali MWEs is presented in section 6. The metrics used for evaluating the systems and experimental results are discussed in section 7. The discussion regarding the utilities and shortcomings of our model is illustrated in section 8 and the concluding part is drawn in the last section.

**1. Multiword Expressions (MWEs)**

Though MWEs are understood quite easily and their acquisition presents no difficulty to native speakers (though it is usually not the case for second language learners), it is hard to identify what features distinguish MWEs from free word combinations. Concerning this issue, the following MWE properties are mentioned in the literature: reduced syntactic and semantic transparency; reduced or lack of compositionality; more or less frozen or fixed status; possible violation of some otherwise general syntactic patterns or rules; a high degree of lexicalization (depending on pragmatic factors); a high degree of conventionality (Calzolari et al., 2002, p. 1934).

No consensus exists so far on the definition of MWEs, but almost all formulations found in research papers emphasize the idiosyncratic nature of this linguistic phenomenon by indicating that MWEs are "*idiosyncratic interpretations that cross word boundaries (or spaces)*" (Sag et al., 2002, p. 1); "*a sequence of words that acts as a single unit at some level of linguistic analysis, ... they are usually instances of well productive syntactic patterns which nevertheless exhibit a peculiar lexical behaviour*" (Calzolari et al., 2002, p. 1934); "*a MWE is composed of two or more words that together form a single unit of meaning, e.g., frying pan, take a stroll, and kick the bucket, … Semantic idiosyncrasy, i.e., the overall meaning of a MWE diverges from the combined contribution of its constituent parts*" (Fazly & Stevenson, 2007, p. 9).

2.1 Noun-Noun MWEs

In the past few years, noun compounds have been a constant source of concern to the researchers towards the goal of full text understanding (Baldwin and Kim, 2010, p. 267 ) (Butnariu et al., 2009, p. 100). Compound nouns are nominal compounds where two or more nouns are combined to form a single phrase such as '*golf club*' or '*computer science department*' (Baldwin et al, 2010, p. 267). There is also a broader class of nominal MWEs where the modifiers are not restricted to be nominal, but can also be verbs (e.g., *hired help*) or adjectives (e.g., *open secret*). To avoid confusion in this article, we will use the term compound nouns when referring to this broader class, throughout the paper, we term this broader class.

Compound noun MWEs can be defined as a lexical unit made up of two or more elements, each of which can function as a lexeme independent of the other(s) when they occur separately in different contexts of the document. The combination of these constituents shows some phonological and/or grammatical isolation from their normal syntactic usages. One property of compound noun MWEs is their underspecified semantics. For example, while sharing the same "head noun" (i.e., rightmost noun in the noun compound), there is less semantic commonality between the components such as *'nut tree'*, *'cloths tree'* and *'family tree'* (Baldwin et al., 2010, p. 267). In each case, the meaning of the compound nouns relates to a sense of both the head and the modifier, but the precise relationship is highly varied and not represented explicitly in any way. Noun-Noun (NN) compounds are the subset of the compound nouns consisting of two consecutive nouns side by side. In English, NN compounds occur in general with high frequency and high lexical and semantic variability. A summary examination of the 90 million word written component of the British National Corpus unearthed over 400,000 NN compound types, with a combined token frequency of 1.3 million; that is, over 1% of words in the BNC are NN compounds (Tanaka and Baldwin, 2003, p. 17).

In Bengali, similar observations are noticed when dealing with the various types of multiword expressions like **compound nouns** (*taser ghar*, 'house of cards', 'fragile'), **complex predicates** such as **conjunct verbs** (*anuvab kara*, 'to feel') and **compound verbs** (*uthe para*, 'to arise'), **idioms** (*matir manus*, 'down to the earth'), **named-entities** (*Rabindranath Thakur,* 'Rabindranath Tagore') etc. Bengali is a language consisting of high morpho-syntactic variation at the surface level. The use of NN multiword expressions in Bengali is quite common. For example, NN compounds especially, idioms (*taser ghar,* 'fragile'), institutionalized phrases (*ranna ghar,* 'kitchen'), named-entities (*Rabindranath Thakur,* 'Rabindranath Tagore'), numbers (*panchso noi,* 'five hundred and nine'), kin terms (*pistuto bh*ai, 'maternal cousin') etc. are very frequently used in Bengali literature. In the next subsection, we classify the compound nouns occurred in Bengali based on their morpho-syntactic properties.

2.2 Classifications of Bengali Compound Noun MWEs

Compound noun MWEs can occur in open (components are separated by space(s)), closed (components are melded together) or hyphenated forms (components are separated by hyphen(s)), and satisfy semantic non-compositionality, statistical co-occurrence or literal phenomena (Kunchukuttan and Damani, 2008, p. 15) etc. Agarwal et al. (2004, p. 165) classified the Bengali MWEs in three main classes using subclasses. Instead, we propose seven broad classes of Bengali compound noun MWEs considering their morpho-syntactic flexibilities, as follows:

- **Named-Entities (NE):** Names of people (*Rabindranath Thakur*, 'Rabindranath Tagore'), names of locations (*Bharat-barsa*, 'India'), names of organizations (*Pashchim Banga Siksha Samsad,* 'West Bengal Board of Education') etc. where inflection is only allowed to be added to the last word.
- **Idiomatic Compound Nouns:** These are non-productive[1] and idiomatic in nature, and inflection can be added only to the last word. The formation of this type is due to the hidden conjunction between the components or absence of inflection from the first component (*maa-baba,* 'mother and father').
- **Idioms:** They are also compound nouns with idiosyncratic meaning, but the first noun is generally in the possessive form (*taser ghar*, 'fragile'). Sometimes, individual components may not carry any significant meaning and may not represent a valid word (*gadai laskari chal,* 'indolent habit'). For them, no inflection is allowed even to the last word.
- **Numbers:** They are highly productive, impenetrable and allow slight syntactic variations like inflections. Inflection can be added only to the last component (*soya sat ghanta,* 'seven hours and fifteen minutes').
- **Relational Noun Compounds**: They are mainly kin terms and consist mostly of two tokens. Inflection can be added to the last word (*pistuto bhai,* 'maternal cousin').
- **Conventionalized Phrases:** Sometimes, they are called as '*Institutionalized phrases*'. Although not necessarily idiomatic, a particular word combination coming to be used to refer to a given object. They are productive and have unexpectedly low frequency and in doing so, contrastively highlight the statistical idiomaticity of the target expression (*bibhha barshiki,* 'marriage anniversary').

---

[1]A phrase is said to be "*productive*" if new phrases can be formed from the combinations of syntactically and semantically similar component words of the original phrase.

Identifying Bengali MWEs using Semantic Clustering    377- **Simile Terms:** They are analogy term in Bengali and sometime similar to the idioms except that they are semi-productive (*hater panch,* 'remaining resource').
- **Reduplicated Terms:** Reduplications are non-productive and tagged as noun phrases. They are further classified as onomatopoeic expressions (*khat khat,* 'knocking'), complete reduplication (*bara-bara,* 'big big'), partial reduplication *(thakur-thukur,* 'God'*),* semantic reduplication (*matha-mundu,* 'head'*),* correlative reduplication (*maramari,* 'fighting') (Chakraborty and Bandyopadhyay, 2010, pp. 72).

Identification of reduplication has already been carried out using the clues of Bengali morphological patterns (Chakraborty and Bandyopadhyay, 2010, pp. 72). A number of research activities in Bengali Named Entity (NE) detection have been conducted (Ekbal et al, 2008, p. 67), but the lack of publicly available standard tools to detect NEs inhibits the incorporation of them within the existing system. Therefore, we discard the identification of NEs from this experiment. Kin terms and numbers can be easily captured by some well-developed lexicons because they are small in number and form a closed set in Bengali (Agarwal et al, 2004, p. 165). The present work mainly focuses on the extraction of productive and semi-productive bigrams, compound noun MWEs like idioms, idiomatic compound nouns, and simile terms (which are in open or hyphenated form) from a document using a semantic clustering technique.

## 3. Semi-automated Approach for Candidate Extraction

### 3.1 Corpus Acquisition and Bigram Extraction

Resource acquisition is one of the challenging obstacles to work with electronically resource constrained languages like Bengali. However, we crawled a large number of Bengali articles written by the noted Indian Nobel laureate Rabindranath Tagore[2]. While we are primarily interested in token level and phrase level characteristics, the information of the document such as the order of the documents, variation of the size of the documents, length normalization etc. has not been maintained and manipulated in the experiment. Therefore, we merged all the articles and prepared a raw corpus consisting of 393,985 tokens and 283,533 types. The actual motivation for choosing the literature domain in the present task was to obtain useful statistics to further help Stylometry analysis (Chakraborty, 2012, p. 41). However in literature, the use of MWEs is greater than in the other domains like tourism, newspapers, scientific documents etc. because the semantic variability of MWEs offers writers more expressive terms. In Bengali literature, idiomatic expressions and relations terms are quite frequently used.

---

[2] http://www.rabindra-rachanabali.nltr.org

Since the preliminary crawled corpus was noisy and unformatted, we used a basic semi-automatic pre-processing technique to make the corpus suitable for parsing. We used a Bengali shallow parser[3] to identify the POS, chunk, root, inflection and other morphological information of each token. We observed that some of the tokens were misspelled due to typographic and phonetic errors. For instance, the token *'boi'* (book) could be written as 'বই' or 'বৈ'. Thus, the Shallow parser could not be able to detect the actual root and inflection of these two variations. To make the system fully automated, we allowed retaining the types of variations into the cleaned text.

After pre-processing, bigram noun sequences whose constituents were in the same chunk were extracted using their POS and chunk categories. We observed that during the parsing phase, the Shallow parser could not disambiguate common nouns ('NN') and proper nouns ('NNP') appropriately. The reason could be the continuous need to coin new terms for new concepts. We took both of them and manually filtered the named-entities from the collected list so that we could accumulate most of the proper nouns for our main experimental module. Although the chunk information helps to identify the boundary of a phrase, some of the phrases belong to chunks having more than two nouns. The frequency of these phrases is also identified during the evaluation phase. Now, a bigram nominal candidate phrase can be thought of as $<M_1\ M_2>$. The morphological heuristics used to separate the candidates are described in Table 1. After the first phase, a list of possible candidates was collected which was fed into the annotation phase.

| | | **Heuristics** |
|---|---|---|
| 1. | POS | POS of each bigram must be either 'NN' or 'NNP' |
| 2. | Chunk | $M_1$ and $M_2$ must be in the same 'NP' chunk |
| 3. | Inflection | Inflection[4] of $M_1$ must be '-শুন্য'(null), '-র'(-r), '-এর'(-er), '-এ'(-e), '-য়'(-y) or '-য়ের'(-yr) and for $M_2$, any inflection is considered |

---

[3] http://ltrc.iiit.ac.in/analyzer/bengali
[4] Chattopadhyay (1992) exlpained that for compound noun MWEs, considerable inflections of the first noun can be those which are mentioned in the table.

*Table 1: Heuristics for the candidate selection*

3.2 Annotation Study

Three anonymous annotators -- linguistic experts working on our project -- were hired to carry out the annotation. They were asked to divide all extracted phrases into four classes and definitions of the classes using the following definitions:
**Class 1: Valid NN MWEs (M)**: phrases which show total non-compositionality and their meanings are hard to predict from their constituents; e.g., হাতের পাঁচ (*hater panch*, 'remaining resource').
**Class 2: Valid NN semantic collocations but not MWEs (S):** phrases which exhibit partial or total compositionality (e.g., act as institutionalized phrases) and show statistical idiomaticity; e.g., বিবাহ বাষিকী (*bibaha barsiki,* 'marriage anniversary').
**Class 3: Invalid collocations (B):** phrases enlisted due to bigrams in an n-gram chunk having more than two components; e.g., পর্বত শহরের (*porbot sohorer,* 'of mountain town').
**Class 4: Invalid candidates (E):** phrases enlisted due to the error in parsing like POS, chunk, inflection including named-entities; e.g., গ্রন্থাগারটি তৈরি (*granthagar tairi,* 'build library').

Class 3 and class 4 types were filtered initially and their individual frequencies are noted as 24.37% and 29.53% respectively. Then the remaining 46.10% (628 phrases) of the total candidates were annotated and labeled as MWE (M) or S (Semantically collocated phrases), and they were fed into the evaluation phase. We plan to make the dataset publicly available soon.
The annotation agreement was measured using standard Cohen's *kappa* coefficient (κ) (Cohen, 1960, p. 37). It is a statistical measure of inter-annotation agreement for qualitative (categorical) items. It measures the agreement between two annotaters who separately classify items into some mutually exclusive categories. We employ another strategy in addition with kappa (κ) to calculate the agreement between annotators. We choose the measure of agreement on set-valued items (MASI) (Passonneau, 2006) that is used for measuring agreement in the semantic and pragmatic annotations. MASI is a distance between sets whose value is 1 for identical sets, and 0 for disjoint sets. For sets A and B, it is defined as: MASI = J * M, where the Jaccard metric (J) is:

$$J = \frac{A \cap B}{A \cup B} \tag{1}$$

Monotonicity (*M*) is defined as follows:

$$M = \begin{cases} 1, & \text{if } A = B \\ 2/3, & \text{if } A \subset B \text{ or } B \subset A \\ 1/3, & \text{if } A \cap B \neq \phi, A - B \neq \phi \text{ and } B - A \neq \phi \\ 0, & \text{if } A \cap B = \phi \end{cases}$$

The inter-annotation agreement scores of three annotators are presented in Table 2. Among the 628 types of noun-noun candidates, half of them selected randomly were used in the development phase and the remaining were used in the testing phase.

| MWEs [# 628] | Agreement between pair of annotators | | | |
|---|---|---|---|---|
| | A1-A2 | A2-A3 | A1-A3 | Average |
| KAPPA | 87.23 | 86.14 | 88.78 | 87.38 |
| MASI | 87.17 | 87.02 | 89.02 | 87.73 |

*Table 2: Inter-annotation agreement*

## 4. Baseline System

As mentioned earlier, the task of identifying Bengali compound nouns from a document has had little attention in the literature, and thus there is no prior developed methodology that can be used for the baseline. Therefore, in this experiment, we simply adapt a heuristic to develop our baseline system. The phrases which do not affix any nominal chunk and determinant at the prefix and suffix positions are selected as MWEs in the baseline system. The baseline system naturally reaches high accuracy in terms of recall since most of the identified MWEs satisfy the heuristics mentioned above. But in terms of precision, it shows very low accuracy (38.68%) since many collocated and fully-compositional elements were wrongly identified as MWEs. The main challenge of our model was to filter these irrelevant collocations from the selected candidate set.

## 5. Statistical Methodologies

We started our experiment with the traditional methodology of collocation detection. Previous literature (Church and Hans, 1990, p. 22; Dunning, 1993, p. 61; Kilgarriff and Rosenzweig, 200, p. 15) shows that various statistical methodologies could be incorporated in identifying MWEs from a large corpus. In this experiment, we developed a statistical system using these previous techniques and modified them according to our requirements[5]. It is worth noting that frequency information of the candidate phrases in a corpus is a strong clue for labeling them as MWEs since it provides the evidence of more certainty of occurrence than randomness. However, for a resource-constrained language like Bengali, infrequent occurrence of candidates may not give any reasonable conclusion to judge them as MWEs (or Non-MWEs) because the size of the corpus itself is generally not adequate for statistical analysis. Therefore, instead of taking the frequency information directly, we took five standard association measures namely Point-wise Mutual Information (PMI) (Church and Hans, 1990, p. 22), Log-Likelihood ratio (LLR) (Dunning, 1993, p. 61), Co-occurrence measure (Agarwal et al, 2004, p. 165), Phi-coefficient and Significance function (Agarwal et al., 2004, p. 165) for extracting NN Multiword Expressions. A combined weighted measurement is proposed for the identification task, which is helpful to compute bigram collocation statistics. We ranked the list individually based on each of the statistical measures. We noticed in the comparative study that the results obtained by the frequency-based statistics like PMI and LLR could not identify MWEs at the top position of the ranked list. Therefore, we posited that the lexico-semantic affinity among the constituents could unleash the dependency of frequency information in the measurement.

Final evaluation combined all the statistical features mentioned above. Experimental results on the development dataset show that Phi-coefficient, Co-occurrence and Significance functions which are actually based on the principle of collocation produce more accurate results compared to direct frequency-based measurements like LLR, PMI in the higher ranks. So, these three measures are considered in the weighted scheme to assign certain weights to the candidate phrases. After a continuous weight tuning over the development data, the best weights for Co-occurrence, Phi and Significance functions are reported as 0.45, 0.35 and 0.20 respectively for the combined measurement. The individual score of each measure is normalized before assigning weights so that they fall in the range of 0 to 1.

For each measurement, the scores have been sorted in descending order and the total range is divided into five bins (five ranks). Here, Rank 1 signifies higher ranked bin. The intuition is that the more the value of the statistical measure for a

---

[5] Interested readers are encouraged to go through the research dissertation by Chakraborty, 2012 (1).

candidate phrase, the more it behaves like a MWE. The metrics used to evaluate the statistical systems are described below:

**Precision in Rank** *i* (**P$_i$**) = (Number of MWEs present in the *i*th ranked bins) ∕ (total number of candidates in *i*th ranked bins)

**Recall in Rank** *i* (**R$_i$**) = (Number of MWEs present in the *i*th ranked bins) ∕ (total number of MWEs in the documents)

**F-score in Rank** *i* (**F$_i$**) = (2*P$_i$*R$_i$) ∕ (P$_i$+R$_i$)

Table 3 shows the results obtained from five association measures and the combined weighted measures over the test dataset.

| Rank | LLR | | | PMI | | | Co-occurrence | | | Phi Coefficient | | |
|---|---|---|---|---|---|---|---|---|---|---|---|---|
| | P | R | F | P | R | F | P | R | F | P | R | F |
| 1 | 17.5 | 15.3 | 16.3 | 18.0 | 18.2 | 18.0 | 34.0 | 24.0 | 28.1 | 35.7 | 32.2 | 33.8 |
| 2 | 16.0 | 13.7 | 14.7 | 16.5 | 26.9 | 20.4 | 22.6 | 28.9 | 25.3 | 21.9 | 24.6 | 23.1 |
| 3 | 20.3 | 27.8 | 23.4 | 18.8 | 17.5 | 18.1 | 18.5 | 12.6 | 14.9 | 15.9 | 16.5 | 16.1 |
| 4 | 19.7 | 29.6 | 23.6 | 22.2 | 24.0 | 23.0 | 11.2 | 19.9 | 14.3 | 17.8 | 11.4 | 13.8 |
| 5 | 20.7 | 13.6 | 16.4 | 23.9 | 13.4 | 17.1 | 10.6 | 14.6 | 12.2 | 15.7 | 15.3 | 15.4 |

| Rank | Significance | | | Weighted Measure (Co-occurrence + Phi+ Significance | | |
|---|---|---|---|---|---|---|
| | P | R | F | P | R | F |
| 1 | 38.5 | 35.7 | 37.0 | 46.5 | 51.0 | 48.6 |
| 2 | 21.6 | 31.2 | 25.5 | 30.2 | 29.8 | 30.0 |
| 3 | 16.1 | 11.9 | 13.6 | 13.4 | 12.5 | 12.9 |
| 4 | 12.3 | 9.6 | 10.7 | 2.6 | 4.2 | 3.2 |
| 5 | 9.7 | 11.6 | 10.5 | 1.0 | 2.3 | 1.4 |

*Table 3: Performance of all the statistical measures along with the weighted measure*

## 6. Semantic Clustering Approach

Multiword Expressions represent a core semantic issue that can be partially resolved by morphological or statistical clues. However, these clues are not sufficient for the present identification task. Our semantic clustering approach aims to handle this problem semantically. This clustering approach tries to cluster semantically related words present in the document. However, identifying semantically related words for a particular token can be carried out by looking at the surroundings tokens and finding the synonymous entries of the surrounding words within a fixed context window. But in that case, higher number of occurrences of a particular expression is essential because one or few occurrences of a particular word cannot explore its actual meaning. Therefore, in a medium-size corpus, it is hard to extract the cluster of synonyms. Since the electronic resources such as newspapers, weblogs may not be present for all the languages and the presence of frequent MWEs in such contents are rare, we focus on extracting the MWEs only from the medium size crawled corpus. However, semantics of a word may be obtained by analyzing its similarity set called the *synset*. Semantic distance of two tokens in a phrase can be measured by comparing their synsets. Higher value of the similarity between two sets indicates semantic closeness of two tokens to each other.

Let M1 and M2 be two components of a bigram <M1 M2>. For each component of the expression, semantically related words present in the documents are extracted by using the formatted Bengali monolingual dictionary (discussed in Section 6.1) and two separate clusters are formed for two tokens. Intersection of two clusters indicates the commonality of two components present in a bigram. Using these common elements, three similarity measurements are proposed and the results are reported separately in Table 5 later. Finally, based on a predefined threshold, the candidate phrases were labeled as MWE or Non-MWE.

6.1 Restructuring the Bengali Monolingual Dictionary

To the best of our knowledge, no full-fledged WordNet or thesaurus is available in Bengali. In this section, we describe the construction of a Bengali thesaurus that aims not only to develop Bengali WordNet but also to identify the meaning of multiword expressions. Focusing mainly on MWEs, the present natural language resource is being developed from the available Bengali-to-Bengali monolingual dictionary (Samsada Bengali Abhidhana[6]). The monolingual dictionary contains each word with its parts-of-speech (বিশেষ্য-Noun, বিশেষণ-Adjective, সর্বনাম-Pronoun, অব্যয়- Indeclinable, ক্রিয়া-Verb), phonetics and synonym sets. Synonym sets are separated using distinguishable notations based on similar or differential meaning. Synonyms of different sense with respect to a word entry are distinguished by a semicolon (;), and synonyms having same sense are separated by a comma (,). An automatic technique is devised to identify the synsets for a particular word entry based on the clues (, and ;) of similar and differential senses. The symbol *tilde* (~) indicates that the suffix string followed by the *tilde* (~) notation makes another new word concatenating with the original entry word. A snapshot of the modified synset entries of the Bengali word "অংশু" (*Angshu*) is shown in Figure 1. Table 4 shows the frequencies of different synsets according to their part-of-speech.

---

[6] http://dsal.uchicago.edu/dictionaries/biswas-bangala/

| Word Entries | Synset | Noun | Adjective | Pronoun | Indeclinable | Verb |
|---|---|---|---|---|---|---|
| 47949 | 63403 | 28485 | 11023 | 235 | 497 | 1709 |

*Table 4: Frequency information of the synsets with different part-of-speeches*

6.2 Generating Synsets of Nouns

At the beginning of the clustering method, we generate a synonym set for each noun present in the corpus using the modified dictionary. The reasons behind the adaptation of the monolingual dictionary in this experiment are manifold: firstly, the major entries of the dictionary are the synsets of *nouns*; secondly, the closed set of dictionary entries limits the size of synsets of nouns and help us compare

**Dictionary Entry:**
অংশু [ aṃśu ] বি. 1 কিরণ , রশ্মি , প্রভা ;  2 আঁশ , তন্তু , সুতোর সূক্ষ্ম অংশ ।  [সং. অন্শ্+উ] ।  ~ ক বি. বস্ত্র , সূক্ষ্ম বস্ত্র ; রেশম পাট ইত্যাদিতে প্রস্তুত বস্ত্র (চীনাংশুক) ।  ~ জাল বি. কিরণরাশি , কিরণমালা ।  ~ ধর বি. অংশুর ধার ; সূর্য ।  ~ মতী বিণ. (স্ত্রী) কিরণময়ী , জ্যোতির্ময়ী ।  ~ মান (-মত্) 1 সূর্য ;  2 সূর্যবংশীয় সগর রাজার পৌত্র ।  ~ মালা বি. রশ্মিজাল , কিরণমালা ।  ~ মালী (-লিন্) বি. সূর্য ।  ~ ল বিণ. কিরণময় , কিরণবিশিষ্ট ।

**Synsets:**
অংশু    কিরণ/রশ্মি/প্রভা_বি.#25_1_1 আঁশ/তন্তু/সুতোর_সূক্ষ্ম_অংশ_[_সং._অন্শ্+উ]_বি.#25_2_2
অংশুক    বস্ত্র/সূক্ষ্ম_বস্ত্র_বি.#26_1_1 রেশম_পাট_ইত্যাদিতে_প্রস্তুত_বস্ত্র_(চীনাংশুক)_বি.#26_2_2
অংশুজাল কিরণরাশি/কিরণমালা_বি.#27_1_1
অংশুধর অংশুর_ধার_বি.#28_1_1 সূর্য_বি.#28_2_2
অংশুমতী (স্ত্রী)_কিরণময়ী/জ্যোতির্ময়ী_বিণ.#29_1_1
অংশুমান সূর্য_#30_1_1 সূর্যবংশীয়_সগর_রাজার_পৌত্র#30_2_2
অংশুমালা    রশ্মিজাল/কিরণমালা_বি.#31_1_1
অংশুমালী    সূর্য_বি.#32_1_1
অংশুল    কিরণময়/কিরণবিশিষ্ট_বিণ.#33_1_1

*Figure 1: Monolingual dictionary entry and built synsets for the word "অংশু" (Angshu)*

two noun synsets; thirdly, the lack of standard Bengali stemmer, lemmatizer can be handled programmatically using the maximum character matching algorithm (Chakraborty and Bandyopadhyay, 2010, p. 72); finally it would be helpful further to infer an idea about different senses of noun tokens present in the corpus. However, the formatted dictionary can be assumed to be a close set of word entries $W^1$, $W^2$, $W^3$,......,$W^m$ where the synsets of the entries look like:

$$W^1 = n_1^1, n_2^1, n_3^1, \ldots = \{n^1\}$$
$$W^2 = n_1^2, n_2^2, n_3^2, \ldots = \{n^2\}$$
$$W^3 = n_1^3, n_2^3, n_3^3, \ldots = \{n^3\}$$
$$\vdots$$
$$W^m = n_1^m, n_2^m, n_3^m, \ldots = \{n^m\}$$

where $W^1$, $W^2$, ..., $W^m$ are the dictionary entries and $\{n^i\}$ denotes the set of synsets of the entry $W^i$. Now each noun entry identified by the shallow parser in the document is searched in the synset entries of the dictionary for its individual existence with or without inflection. For instance, $N$ is a noun in the corpus and it is present in the synsets of $W^1$, $W^3$ and $W^5$. Therefore, they become entries of the synset of $N$. Formally, this can be represented as follows.

$$Synset(N) = \{W^1, W^3, W^5\} \quad (2)$$

Equation 2 states that since the given noun N is present in the synsets of $W^1$, $W^3$ and $W^5$, the sense of these three dictionary entries are somehow related to the sense of $N$. Following this, the synonym noun tokens for each of the nouns present in the corpus are extracted from the dictionary. In short, the formatted dictionary indeed helps us cluster synonymous tokens corresponding to a particular noun present in a document.

6.3 Semantic Relatedness between Noun Synsets

The next task is to identify the similarity between the synsets of two nouns that can help predict the semantic relatedness between them. This is done by taking the intersection of the synsets and assigning a score to each such noun-pair to indicate the semantic affinity between two nouns. If $N_i$ and $N_j$ are two nouns in the document, and $S_i$ and $S_j$ are their corresponding synsets extracted using the technique stated in Section 6.2, then the commonality of the two nouns can be defined as:

$$Comm(N_i, N_j) = |S_i \cap S_j| \quad (3)$$

The above equation shows that the commonality is maximum when the similarity is measured with itself (i.e., $Comm(N_i, N_j)$ is maximum when $i = j$). This semantic commonality measurement approximately gives a score according to the

386   Chakraborty et al.

commonality of their synset entries. This score helps further to build the cluster of each of the nouns present in the document. This is discussed in the next subsection.

6.4 Semantic Clustering of Nouns

All the tokens in a document identified as nouns by the Shallow parser are tagged as spatial and temporal expression or 'NNP' (proper noun). In this experiment, we mainly used common nouns. Using the scores obtained by the semantic commonality measure discussed in the previous section, we can build a cluster centered on a given noun present in the document such that the cluster constitutes all the nouns semantically related to the given noun (discussed in subsections 6.2 and 6.3). A score is assigned to each such noun present in the cluster representing the semantic similarity (discussed in subsection 6.3) between this noun to the centre noun. For example, suppose the nouns identified by the Shallow parser in the document are $W_1$, $W_2$, …,$W_i$, $W_j$, $W_k$, $W_l$, $W_m$, $W_n$ etc. Now, for a given noun M, the semantic cluster of M and the association scores with the constituent elements of the cluster are shown in Figure 2. In this figure, the semantic similarities of M with the other nouns are denoted by the weights (i.e., a, b, c etc.) of the edges.

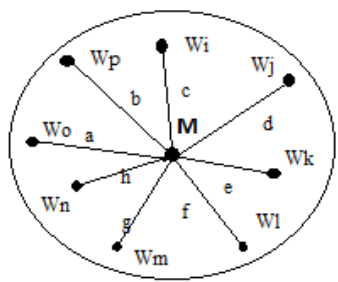

*Figure 2: Semantic clustering of M and the associated commonality scores with similar nouns*

6.5 Decision Algorithm for Identifying MWEs

We extract the candidates eligible for judging MWE in section 3. The elaborated algorithm to identify a noun-noun bigram (say, $< M_1\ M_2>$) as MWEs is discussed below with an illustrative example shown in Figure 3.

Here, we elaborate step 3 and step 4 since the central theme of the algorithm lies in these two steps. After identifying the common terms from the synsets of the components of a candidate, a vector space model is used to identify the similarity between the two components. In *n*-dimensional vector space, these common elements denote the axes and each candidate acts as a point in the *n*-dimensional space. The coordinate position of the point (each component of the candidate bigram) in each direction is represented by the similarity measure between the synsets of each component and the noun representing the axis in that direction. The cut-off value for the classification of a given candidate as MWE (or Non-MWE) is determined from the development dataset after several tries to get the best performance (described in step 4). In the experiment, we observe that the bigrams that are actual MWEs, mainly non-compositional phrases, show a low similarity score between the synsets of their components.

**Algorithm:MWE_CHECKING**
**Input:** Noun-noun bigram < $M_1$ $M_2$>
**Output**: Return true if MWE, or return false.
1. Extract semantic clusters of $M_1$ and $M_2$ (discussed in Section 6.4);
2. Intersect the clusters of $M_1$ and $M_2$ (Figure 3 (left) shows the common synset entries (broken rectangles) of $M_1$ and $M_2$);
3. For measuring the semantic similarity between $M_1$ and $M_2$:
      3.1. In an *n*-dimensional vector space (*n* denotes the number of elements common in both the synsets of $M_1$ and $M_2$, e.g., in the Figure 3 (left), n=2), the common entries act as the axes. Put $M_1$ and $M_2$ as two vectors and their associated similarity with the axes tokens as their co-ordinates.
      3.2. Calculate cosine-similarity measurement and Euclidean distance between the two vectors (Figure 3 (right)).
4. Final decision is taken separately for two different measurements:
      4.1 If (cosine-similarity > α) return false; else return true;
      4.2 If (Euclidean distance > β) return false; else return true;
      (where α and β are the pre-defined cut-off values determined from the development set)

If we take an example of the Bengali idiom -- *hater panch* ('remaining resource'), we can see that English WordNet defines two components of the idiom in the following way: *hat* ('hand') as 'a part of a limb that is farthest from the torso' and *panch* ('five') as 'a number which is one more than four'. So from these two glosses it is quite evident that they are not at all semantically related. The synonym sets for these two components extracted from the formatted monolingual dictionary are as follows.

Synset (হাত) = { হস্ত (hasta), কর (kar), পাণি (pani), বাহু (bahu), ভুজ (bhuj), কৌশল (kaushal), হস্তক্ষেপ (hastakkhep), ধারণ (dharan), রেখা (rekha), লিখিত (likhita), হস্তাক্ষর (hastakshar), হস্তান্তর (hastantar), হাজা (haza)}

Synset (পাঁচ) = {পঞ্চ (pancha), সংখ্যা (sankha), কর্ম (karma), গঙ্গা (ganga), গব্য (gobbo), কন্যা (kannya), গুণ (gunn), গৌড় (gourya), তন্ত্র (tantrya), তীর্থ (tarthya), পঞ্চত্ব (panchanta), পনেরো (ponero), পূর্ণিমা (purnima), পঞ্চাশ (panchas) }

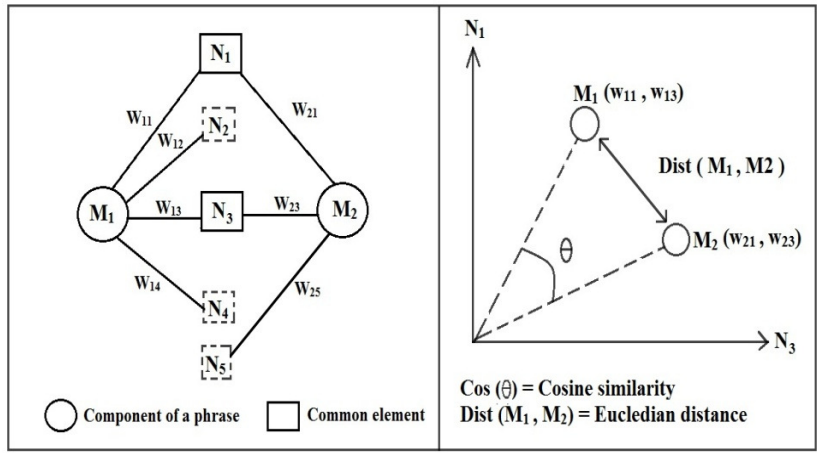

*Figure 3: Intersection between the clusters of the components of a candidate bigram (left) and the similarity between two components (right)*

We can observe that the two synonym sets have no element in common and therefore their similarity score would be zero. In this case, a vector space model cannot be drawn in zero dimensional space. For them, a final concession weight is assigned to treat them as fully non-compositional phrases. To identify their non-compositionality, we need to show that their occurrences are not by mistake (i.e., because of a typo or due to unawareness of the author); rather they can occur side by side in several instances. But the concrete statistical proof can only be obtained using a large corpus. Here, for the candidate phrases which have zero similarity, we observe their existence more than one time in the corpus and then treat them as MWEs.

## 7. WordNet::Similarity Measurement

We also incorporate English WordNet 2.17 in this experiment to measure the semantic distance between two Bengali words after translating them into English. Though the idea is trivial considering the manual intervention of the translation process, our main focus was to get an idea of how the semantic similarity of two components can help identify the combination as an MWE, and how a well-defined lexical tool is essential in the presently adapted linguistic environment. As already mentioned, WordNet::Similarity is an open-source package developed at the University of Minnesota for calculating the lexical similarity between word (or sense). Basically, it provides six measures of similarity and three measures of relatedness based on the WordNet lexical database (Fellbaum, 1998). The measures are based on the analysis of the WordNet hierarchy.

The measures of similarity are divided into two groups: path-based and information content-based. We chose two similarity measures in WordNet::Similarity for our experiments: WUP and LCH; WUP finds the path length to the root node from the least common subsumer (LCS) of the two word senses that is the most specific word sense they share as an ancestor (Wu and Palmar, 1994, p. 133) by the Equation 4.

$$Norm\_dist = \frac{Min\_Dist\_CP}{Dist\_CP\_to\_R + Min\_dist\_CP} \quad (4)$$

Where Norm_dist = calculated normalized distance; Min_Dist_CP = minimum distance to common parent; Dist_CP_to_R= distance from common parent to the root.

In this experiment, we first translate the root of two Bengali components in a candidate phrase into their English forms using the Bengali-to-English Bilingual Dictionary[8]. Then these two words are run through the WordNet based Similarity module for measuring their semantic distance. A predefined cut-off value (μ) is determined from the development set to distinguish between an MWE and a simple compositional term. If the measured distance is less than the threshold, the similarity between them is less. The results are noted for different cut-off values as shown in Table 5. The bold font in each column shows the highest accuracy among different cut-off values.

---

[7] http://www.d.umn.edu/~tpederse/similarity.html
[8] http://dsal.uchicago.edu/dictionaries/biswas-bengali/

| Cut-off | Cosine-Similarity | | | Euclidean Distance | | | WordNet::Similarity | | |
|---|---|---|---|---|---|---|---|---|---|
| | P | R | F | P | R | F | P | R | F |
| 0.6 | 70.75 | **64.87** | 67.68 | 70.57 | **62.23** | 66.14 | 74.60 | **61.78** | 67.58 |
| 0.5 | **78.56** | 59.45 | **67.74** | 72.97 | 58.79 | 65.12 | **80.90** | 58.75 | **68.06** |
| 0.4 | 73.23 | 56.97 | 64.08 | **79.78** | 53.03 | 63.71 | 75.09 | 52.27 | **61.63** |

*Table 5: Precision, Recall and F-score of three measures (in %) in clustering approach and WordNet::Similarity measure*

## 9. Discussion

At the beginning of the article, we claimed that the increasing degree of semantic similarity between two constituents of a candidate bigram indicates the less probability of the candidate to be a multiword expression. The statistical methodologies achieve low accuracy because the medium size corpus fails to unfold significant clue of their occurrences to label the non-compositional phrase as MWEs. We have adopted an approach taking into account the semantic interpretation of MWE that seems to be unconventional in the task of identifying MWEs in any language. In the experimental results, the semantic clustering approach outperforms the other systems. However, the clustering algorithm is able to identify those MWEs whose semantics are fully opaque from the semantics of their constituents (strictly non-compositional). But MWEs show a continuum spectrum from fully-compositional (e.g., idioms) to institutionalized phrases (*e.g., traffic signal*) where high statistical occurrence is the only clue to identify them as MWEs. These partial or transparent expressions are not captured by our system because of the lack of a large size standard corpus.

The presence of the monolingual dictionary is another important criterion to carry out the proposed approach. It acts as a proxy for an individual noun to cumulate the related noun tokens. This algorithm assumes that every language should possess its own dictionary since it is the first and fundamental resource used not only for experimental purposes but also for language generation and understanding.

## Conclusion

We hypothesized that sense induction using synonym set can assist in identifying multiword expressions in Bengali. We introduced a semi-automated approach to establish the hypothesis. We compared our results with the baseline system and the traditional statistical systems. We have shown that clustering measure can be an effective measure to enhance the extraction task of MWEs. The contributions of the paper are fourfold: firstly, we provide an efficient way of clustering noun tokens having similar sense; secondly, we propose a semantic similarity based approach for identifying MWEs; thirdly, it a preliminary attempt to reconstruct a Bengali monolingual dictionary as a standard lexical thesaurus and finally, the present task is a pioneering step towards the development of Bengali WordNet. At last, we would like to stress that this entire methodology can be used to identify MWEs in any other language domain.

In the future, we plan to extend the algorithm to support all ranges of compositionality of Bengali MWEs. Moreover, we modify the semantic interpretation of MWEs to enlist partial and compositional phrases as much as possible. Furthermore, incorporating the Named-Entity recognizer can help develop a full-fledged MWE identification system. Finally, we will make the formatted monolingual dictionary publicly available soon and incorporate the strictly non-compositional MWEs which rarely occur in the medium-size corpus into the dictionary so that they are directly captured from the thesaurus.

**Abstract**

*Identifying Bengali Multiword Expressions using Semantic Clustering*

One of the key issues in both natural language understanding and generation is the appropriate processing of Multiword Expressions (MWEs). MWEs pose a huge problem to the precise language processing due to their idiosyncratic nature and diversity in lexical, syntactical and semantic properties. The semantics of a MWE cannot be expressed after combining the semantics of its constituents. Therefore, the formalism of semantic clustering is often viewed as an instrument for extracting MWEs especially for resource constraint languages like Bengali. The present semantic clustering approach contributes to locate clusters of the synonymous noun tokens present in the document. These clusters in turn help measure the similarity between the constituent words of a potentially candidate phrase using a vector space model and judge the suitability of this phrase to be a MWE. In this experiment, we apply the semantic clustering approach for noun-noun bigram MWEs, though it can be extended to any types of MWEs. In parallel, the well known statistical models namely Pointwise Mutual Information (PMI), Log Likelihood Ratio (LLR), Significance function are also employed to extract MWEs from the Bengali corpus. The comparative evaluation shows that the semantic clustering approach outperforms all other competing statistical models. As a byproduct of this experiment, we have started developing a standard lexicon in Bengali that serves as a productive Bengali linguistic thesaurus.

**Keywords:** Multiword expressions, collocation, idiom, semantic clustering, Bengali.

*Author's address:*

*Tanmoy Chakraborty*
*Department of Computer Science & Engineering*
*Indian Institute of Technology, Kharagpur, India-721302*
*its_tanmoy@cse.iitkgp.ernet.in; its_tanmoy@yahoo.co.in*
*http://cse.iitkgp.ac.in/~tanmoyc*

*Dipankar Das*
*Department of Computer Science & Engineering*
*National Institute of Technology, Meghalaya, India*
*dkaiser@iicm.edu*
*http://www.dasdipankar.com/*

*Sivaji Bandyopadhyay*
*Department of Computer Science & Engineering*
*Jadavpur University, Kolkata, India*
*sivaji_cse_ju@yahoo.com*
*http://www.sivajibandyopadhyay.com/*